\documentclass[runningheads]{llncs}
\usepackage[T1]{fontenc}
\usepackage{booktabs}
\usepackage{stmaryrd}
\usepackage{amssymb}
\usepackage{amsfonts}
\AtBeginDocument{%
  }

\usepackage{graphicx}
\usepackage{tcolorbox}
\usepackage{dialogue}
\usepackage[skip=5pt]{caption}
\usepackage[capitalise, noabbrev]{cleveref}
\begin{document}

\title{The Impact of Background Speech on Interruption Detection in Collaborative Groups}
\titlerunning{The Impact of Background Speech on Interruption Detection}

\author {Mariah Bradford\orcidID{0009-0009-2162-3307} \and Nikhil Krishnaswamy\orcidID{0000-0001-7878-7227} \and Nathaniel Blanchard\orcidID{0000-0002-2653-0873}}
\authorrunning{M. Bradford et al.}
%
\institute{Colorado State University, Fort Collins, CO 80523, USA \\
\email{mbrad@rams.colostate.edu}
}


\maketitle
\begin{abstract}
Interruption plays a crucial role in collaborative learning, shaping group interactions and influencing knowledge construction. AI-driven support can assist teachers in monitoring these interactions. However, most previous work on interruption detection and interpretation has been conducted in single-conversation environments with relatively clean audio. AI agents deployed in classrooms for collaborative learning within small groups will need to contend with multiple concurrent conversations --- in this context, overlapping speech will be ubiquitous, and interruptions will need to be identified in other ways. In this work, we analyze interruption detection in single-conversation and multi-group dialogue settings. We then create a state-of-the-art method for interruption identification that is robust to overlapping speech, and thus could be deployed in classrooms. Further, our work highlights meaningful linguistic and prosodic information about how interruptions manifest in collaborative group interactions. Our investigation also paves the way for future works to account for the influence of overlapping speech from multiple groups when tracking group dialog. 
\end{abstract}




%

\section{Introduction}

As artificially intelligent (AI) agents become integrated into learning environments, there is growing opportunity for educators to deploy AI tools within small group work. As a classroom splits into many groups, a teacher would benefit from having each group automatically tracked, providing a rich overview of the classroom and easy identification of struggling groups.
However, such an environment would inherently become challenging to agents; with many groups, there is sure to be a high level of noise and overlapping speech. An interesting problem arises from this environment: we can no longer operationalize interruptions as the presence of overlapping speech. As interruptions are known to be an important component of dialogue and collaboration~\cite{sun_towards_2020,yang_visualizing_2003}, it is important to ensure a system that is robust to this problem in noisy speech environments.

Previous work has established the importance of interruptions in collaboration and communication \cite{foroughi_interruptions_2014,peters_when_2017,puranik_pardon_2020,sun_towards_2020,yang_visualizing_2003}. Automatic tracking of interruptions can help teachers moderate many students at once during group work in the classroom \cite{sun_towards_2020}. Tracking can give teachers insight into the speaker balance in the conversation, agreement and disagreement, topic shifts, and engagement \cite{yang_visualizing_2003}. 
In particular, interruptions are important indicators: groups with high levels of interruptions may reveal uncertainty in a group, particularly when members are interrupting frequently to request clarification or to express disagreement, indicating the group is struggling to build common ground~\cite{yang_visualizing_2003}. Conversely, interruptions can also be indicative of engagement in a task~\cite{yang_visualizing_2003}.

\begin{figure*}[t]
\captionbox{Interruption with overlapping speech. Participant 1 (left): "This one is really heavy so let's do this and a ten." Participant 3 (right): "Twenty, thirty, forty."\label{fig:interrupt}} {
  \includegraphics[width=0.48\textwidth]{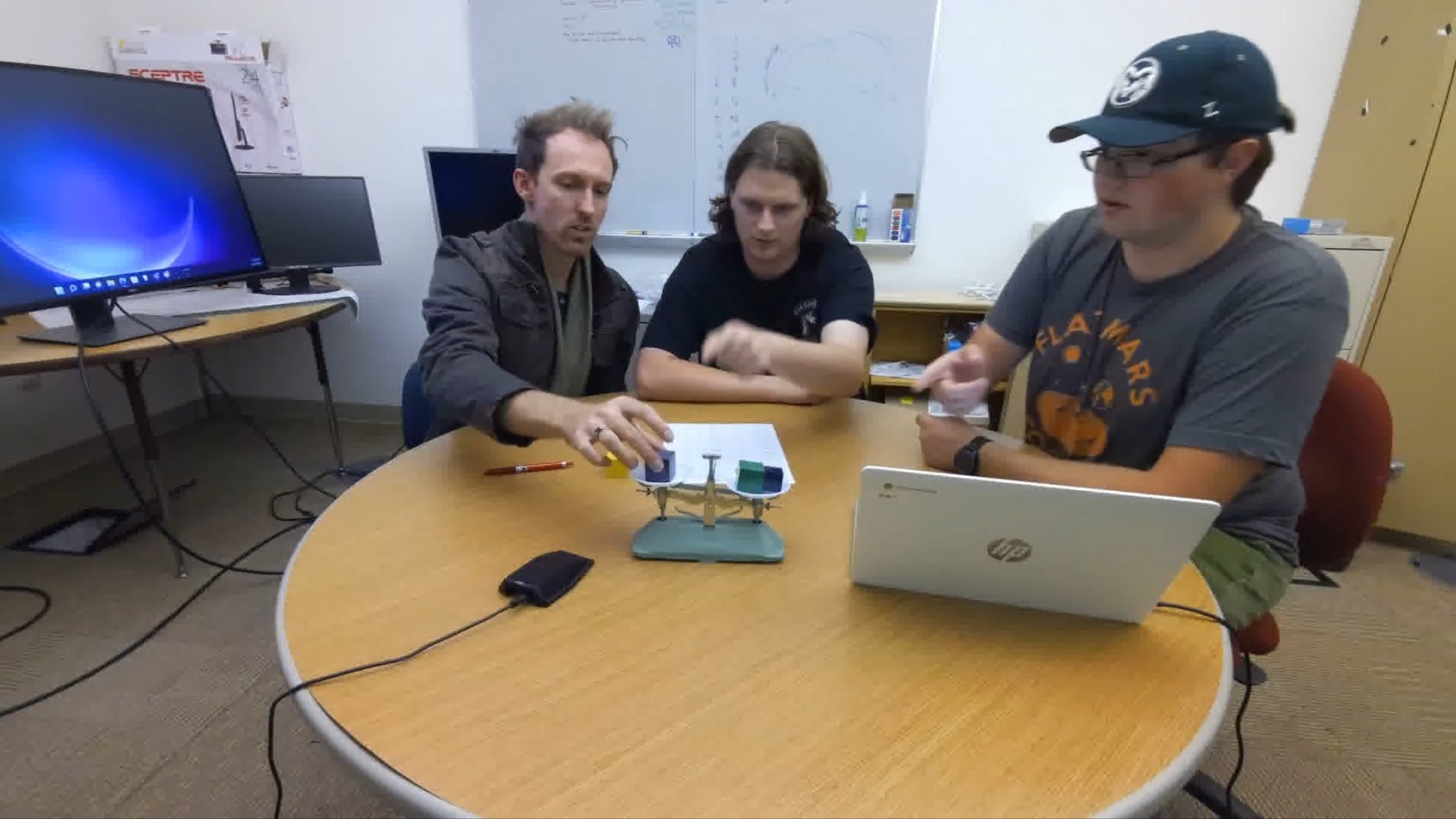}} \hfill
\captionbox{Interruption with no overlapping speech, during a lull. Participant 2 (middle): "Yeah, it's..." Participant 1 (left): "It's pretty far to the right side now." \label{fig:lull}} {
  \includegraphics[width=0.48\textwidth]{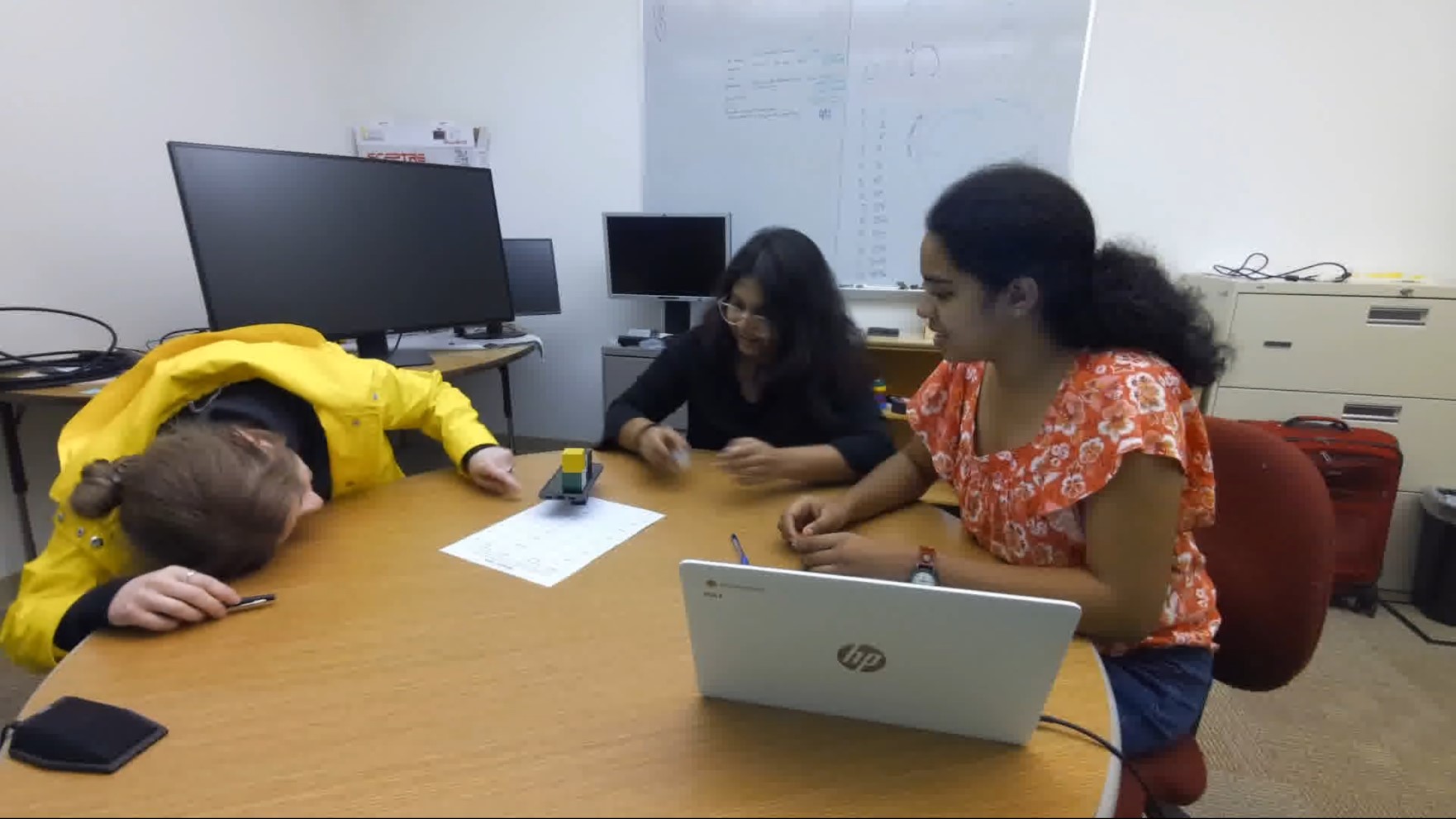}}
\end{figure*}

To track and classify interruptions for small group work in noisy classrooms, an automated system will need a strong feature set. In these scenarios, simple speech-overlap detection will not be enough because there will inherently be many speakers at once. Interruptions happen \textit{within} the group’s conversation, and an agent will need to be robust to the target group's behavior. Additionally, interruptions can occur during lulls in speech, resulting in an interruption without any overlap in speech. An example of this can be seen in Figure \ref{fig:lull}. Therefore, an agent needs a variety of features to build a robust representation of the group and accurately classify interruptions.

In this work, we focus on demonstrating the problem of noisy audio settings. We first detect interruptions using textual and acoustic (including prosodic) features in a clean setting, where only audio from one group exists. We then repeat this process over noisy audio, where the audio stream contains the simultaneous dialogues from multiple groups interacting.

Our research questions are as follows:

\textbf{RQ1:} What features enable accurate detection of interruptions in group conversations?

\textbf{RQ2:} How does background noise from other groups affect the performance of an automated interruption detection system trained on the features identified in RQ1?

\textbf{RQ3:} Which features remain effective for detecting interruptions despite the presence of background noise from other groups?

Our novel contributions include a nuanced exploration of interruption detection. We compare overlapping speech to human-labeled interruptions to evaluate the differences. We evaluate our implementation of interruption detection in a noisy environment. Toward that end, we create overlapping speech data using the original single-group audio. Further, we perform an ablation analysis of feature robustness in interruption detection in noisy settings.

\section{Related Work}

Previous work has established the importance of interruptions in collaboration and communication~\cite{sun_towards_2020,yang_visualizing_2003}. Collaborative problem solving is a common pedagogical tool in the classroom, where students work in small groups to complete a given task \cite{graesser_advancing_2018}. This is both a method of learning and an important skill to develop \cite{oecd_pisa_2017}. Within collaboration, the learning sciences have established interruptions as an important event to track \cite{sun_towards_2020}. 
Yang \cite{yang_visualizing_2003} grouped conversational interruptions into two categories: cooperative and  competitive. The timing and content of interruptions also changes their impact. Foroughi et al. \cite{foroughi_interruptions_2014} found that off-topic interruptions disrupted the quality of work during both planning and execution. In a review of work, Puranik et al. \cite{puranik_pardon_2020} found that some interruptions have a negative impact on productivity and worker well-being; however, they can have positive impacts when they occur during key times. Peters et al. \cite{peters_when_2017} also found that the timing of interruptions impacts the way they affect task performance. With this in mind, Sun et al. \cite{sun_towards_2020} identified interruptions as an important, but reverse-coded, indicator for collaborative problem-solving.

Other works have explored predicting and detecting interruptions and other disfluencies in speech. Lee and Narayanan \cite{lee_predicting_2010} used multimodal cues to predict interruptions in dyadic conversation using a hidden conditional random field and found that both prosodic and gestural features are useful. 
Prior work has also used abrupt changes in acoustic features to identify interruptions and other disfluencies such as stuttering~\cite{biron_automatic_2021,liu_automatic_2023}. 
Current methods explore interruptions with no ambient speech overlap; in this work we argue that these unique conditions warrant exploring interruptions in a multi-group setting.


\section{Methodology}

\subsection{Dataset}
\label{sec:WTD}
We evaluate interruption detection in small groups on the Weights Task Dataset (WTD; \cite{khebour2024text}). This dataset includes audiovisual recordings from three fixed camera views of ten triads solving a puzzle with blocks and a scale. The groups use the scale and combinations of blocks to identify the weights of all blocks and must ultimately discover the pattern inherent in the block weights (an instance of the Fibonacci sequence). Participants communicate using multiple modalities and the dataset is labeled with multiple types of moves (or "indicators") made in the course of collaborative problem solving (CPS) dialogues. The CPS indicators were drawn from the framework defined by Sun et al.~\cite{sun_towards_2020}, which includes 19 indicators important to CPS, including labels for participant interruptions. This allows us to explore natural interruptions within small group work through the lens of multiple modalities---specifically, we examine transcribed text and acoustic features from this dataset to predict the interruption labels. To our knowledge, no other study has used this dataset to develop interruption-detection models.

The Weights Task Dataset is annotated with utterances segmented and transcribed by human annotators (termed "Oracle," being the gold standard), and by Google Automatic Speech Recognition, using the Google Voice Activity Detector for segmentation. Terpstra et al. \cite{terpstra2023good} already examined the effects of Oracle vs. automatic segmentation on CPS annotations over this dataset. Because we are concerned primarily with replicating the conditions of an automatic agent, we do not experiment with Oracle segments or transcripts, but reserve them as an analysis tool (see Section~\ref{sec:disc}).

Figure~\ref{fig:interrupt} shows an example still from the WTD at the moment of an interruption that consists of overlapping speech, with the accompanying utterances. Figure~\ref{fig:lull}, by contrast, shows an example of an interruption that does not correspond to overlapping speech. In this example, Participant 2 has begun but not completed their sentence when Participant 1 interrupts. This example and the evident diversity of circumstances under which interruptions may occur demonstrate how treating interruptions as synonymous with overlapping speech is insufficient in realistic scenarios.

\subsubsection{Multi-group recordings}
\label{sssec:noisy}

For this study, we used audio recordings of the same groups collaborating in the WTD. We used spatial effects to simulate how sound travels in a physical space when overlaying these groups' audio files in order to approximated a shared acoustic environment where multiple groups progressed through the task at once. Using this approach, we were able to generate data for multiple combinations of groups including if only one group was present, if two were, etc. all the way to all 10 groups progressing through the task at once. 

\subsection{Feature Extraction}
\label{feature extraction}
We followed the methods described in~\cite{bradford_automatic_2023} to automatically segment and extract features from this dataset. Beginning when participants started the task, a continuous audio file featuring all participants' speech was recorded. Subsequently, utterances were automatically extracted from this file using Google's Voice Activity Detector (VAD) \cite{karrer_google_2022}. This tool took in an audio stream and segments it into individual utterances, delimited by silence.  
Because utterances are not always cleanly separated at the individual speaker level, an utterance may contain speech from multiple speakers, including overlapping speech.

Across all 10 groups, clean audio was automatically segmented into 1,822 utterances averaging 4.26 (\textit{SD} = 2.85) seconds each. We repeated this process with the noisy audio files and segmented 798 utterances, averaging 12.45 (\textit{SD} = 11.63) seconds. Additionally, we retained the timestamps from the clean audio segmentation and applied them to the noisy audio in order to explore interruptions while controlling for the behavior of the segmentation tool.

\subsubsection{Transcribed Speech Features} We used Google automatic speech recognition (ASR) to transcribe each utterance. Then, we used the BERT-small model~\cite{turc_well_2019} to retrieve word embeddings for the transcriptions. The preprocessing phase before retrieving the BERT-small embeddings included removing special characters and adding the necessary {\tt [CLS]} and {\tt [SEP]} tokens. Each transcription resulted in a 512-dimensional feature vector. 

\subsubsection{Acoustic Features} We used openSMILE \cite{eyben2010opensmile} to extract acoustic features from each utterance. This software allows us to extract a wide range of acoustic features from audio. We narrowed down the available features to a select few by using the extended Geneva Minimalistic Acoustic Parameter Set (eGeMAPS)~\cite{eyben_geneva_2016}. This feature set is a minimal set of 88 features that captures critical acoustic information such as features relevant to frequency, energy, and balance, which is ideal for the eventual task of real-time feature extraction. We extracted one set of acoustic features for every segment.

\subsection{Labels}
\label{sec:labels}
We assigned the interruption labels to the corresponding utterances (described in Section~\ref{feature extraction}) by aligning the timestamps of the CPS indicators from the dataset with those of utterances. If an interruption was found to occur at any point in the utterance, the utterance was labeled as containing an interruption. To compare interruption detection to detection of other similarly-occurring CPS indicators in the dataset, we also retrieved labels for the other three most frequently occurring indicators: \textit{confirming understanding}, \textit{monitoring results}, and \textit{responding to others}. These four labels were the most common in the dataset. Including additional labels enables us to assess how noise specifically affects interruption detection compared to detection of other indicators.

\subsection{Model Search}
We compared the following methods in our model search: Adaboost, Naive Bayes, Logistic Regression, Support Vector Classifier, Decision Tree, Random Forest, and K-Nearest Neighbors. We used the Hyperopt~\cite{bergstra_making_2013} to conduct a guided model search to identify the best model and hyperparameters using F1 score as the guiding metric. Given there are only ten groups in the dataset, we used a leave-one-group-out method for cross validation. 
Similarly to \cite{bradford_automatic_2023}, where a random forest classifier was the best performing model on CPS facet classification, we also found random forest classifiers to perform best in this binary classification task. We report our findings from our random forest models.

\subsection{Experiments}
We conducted three experiments. {\bf Experiment 1} was a replication of results over the original dataset with {\it clean} audio data (no background noise and segmentation run over clean audio as well). 
The other two experiments used the noisy data. {\bf Experiment 2} was to apply the same pipeline to noisy data to explore the impact of noise on interruption detection, assuming that noise would also impact segmentation in a real scenario. This is evaluated using length of segments and F1 score, and exposes where noisy audio affects automatic segmentation, which in turn affects interruption detection performed over those segments.
{\bf Experiment 3} held the segmentation method constant by applying the clean audio timestamps to the noisy audio. This allowed us to explore the robustness of individual feature types when detecting interruptions under noisy conditions. Additionally, all experiments were conducted over the other three labels (\textit{confirming understanding}, \textit{monitoring results}, and \textit{responding to others}) for further comparison. 

\section{Results}
\label{sec:results}

\subsubsection{Baseline}
\begin{figure}
\centering
  \includegraphics[width=0.4\textwidth]{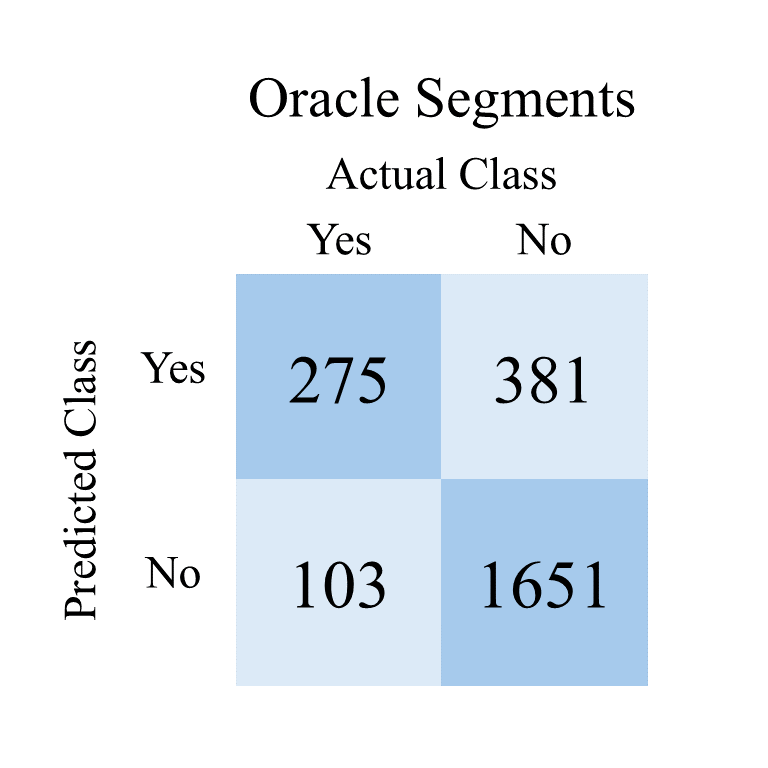}
  \caption{Confusion matrix of the heuristic baseline when using overlap as the predictor of interruptions. The number of false positives for this naive "classifier" underscores the issues with using overlapping speech as a predictor of interruptions. Further, there are 103 interruptions that do not manifest with any overlapping speech, highlighting that interruptions need to be identified with linguistic and prosodic indicators.}
  \label{fig:oracle_cm}
\end{figure}

\begin{figure}
\centering
  \includegraphics[width=.8\columnwidth]{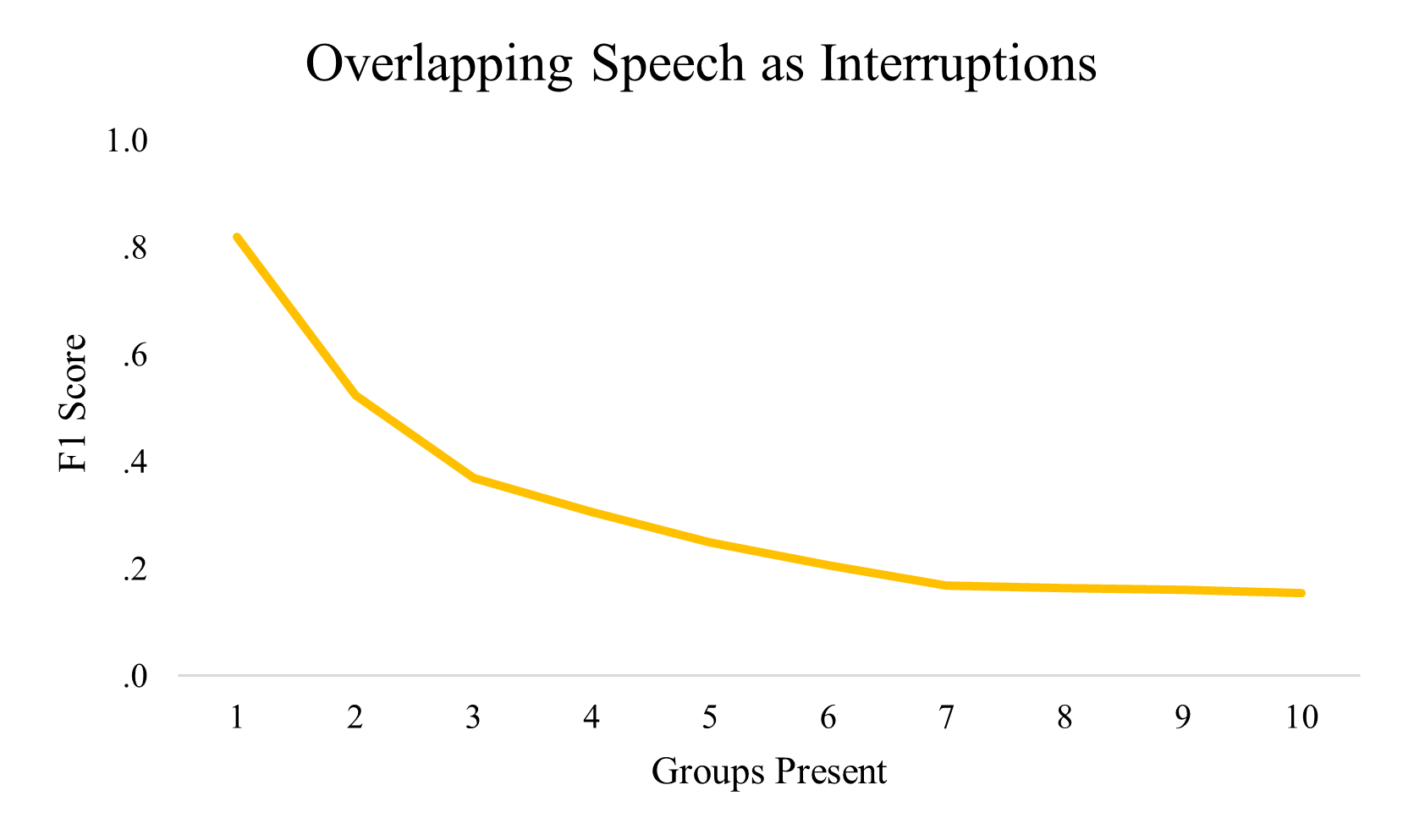}
  \caption{F1 degradation of the heuristic baseline as the number of groups (and overlapping speech) increases. }
  \label{fig:overlap}
\end{figure}

To replicate performance assuming a definition of interruptions as synonymous with overlapping speech, we calculated a heuristic baseline in analogues of the "clean" and "noisy" conditions. The heuristic extracted utterances whose Oracle timestamps overlapped with the Oracle timestamps of another utterance which had started first. By labeling all such utterances as interruptions and comparing them to utterances actually {\it labeled} as interruptions, we calculated how interruption detection would perform assuming the "overlapping speech" definition. For an analogue to clean data, we conducted this evaluation only over utterances within each group. A confusion matrix for this comparison can be seen in Figure~\ref{fig:oracle_cm}. In the noisy analogue, we aligned the utterance timestamps of all 10 groups to the same starting point and if any two utterances overlapped, regardless of which group they came from, they were labeled as interruptions.
When using this method to label interruptions, in the clean analogue, we achieve an F1 score of .819, which is similar to results seen in Table~\ref{tab:f1-clean-clean}. When we move into the noisy analogue, that score drops dramatically to .524, demonstrating that in a noisy environment, assuming that interruptions correspond to overlapping speech is far less useful. Figure~\ref{fig:overlap} shows the degradation of performance as we progressively include more groups in this "overlap space". Sounds from surrounding groups rapidly degrade interruptions operationalized as overlapping speech, as shown by the drop in F1 in this graph. In this work, we show our methodology for interruption detection is robust to such noise.

Using F1 score as the primary metric, we present results from the three experiments described above, using combined transcribed speech and acoustic features, and ablating the contribution of each feature type to each indicator. The numbers presented are weighted by label and averaged across groups. We also present a baseline discussed below. Further, we generated random predictions for the interruption label and we estimate a chance F1 score to be .567 (\textit{SD} = .009) for the oracle segments, .562 (\textit{SD} = .011) for the clean segments, and .523 (\textit{SD} = .017) for the noisy segments.

\subsubsection{Experiment 1: Clean Audio, Clean Segments}
\label{ssec:clean-clean}

Table~\ref{tab:f1-clean-clean} shows results of interruption detection and other included labels in the default condition (single-group clean audio with clean audio segmentation). Performance on detecting interruptions is better than detecting the other comparable indicators. Acoustic features led to higher interruption detection performance compared with textual features --- textual features led to higher performance for all other indicators. 
\begin{table*}[ht!]
\centering
\caption{Average F1 score (standard deviation in parenthesis) using clean segments and clean audio.}
\setlength{\tabcolsep}{6pt} 
\begin{tabular}{rcccc}
\toprule
& \textbf{Interrupts} & \textbf{\begin{tabular}[c]{@{}l@{}}Confirms\\ Understanding\end{tabular}} & \textbf{\begin{tabular}[c]{@{}l@{}}Monitors\\ Results\end{tabular}} & \textbf{Responds} \\ \midrule
\textbf{Textual}     & .798 (.055) & .792 (.052) & .768 (.059) & .781 (.053) \\
\textbf{Acoustic}   & .821 (.029) & .764 (.057) & .731 (.068) & .776 (.050) \\
\textbf{Textual+Acoustic} & .819 (.032) & .795 (.051) & .771 (.054) & .781 (.043) \\
\hline
\end{tabular}
\label{tab:f1-clean-clean}
\end{table*}

\subsubsection{Experiment 2: Noisy Audio, Noisy Segments}
\label{ssec:exp2}

\begin{figure*}
\captionbox{Confusion matrix of the prediction model using clean segments and clean audio.\label{fig:clean_cm}} {
  \includegraphics[width=0.4\textwidth]{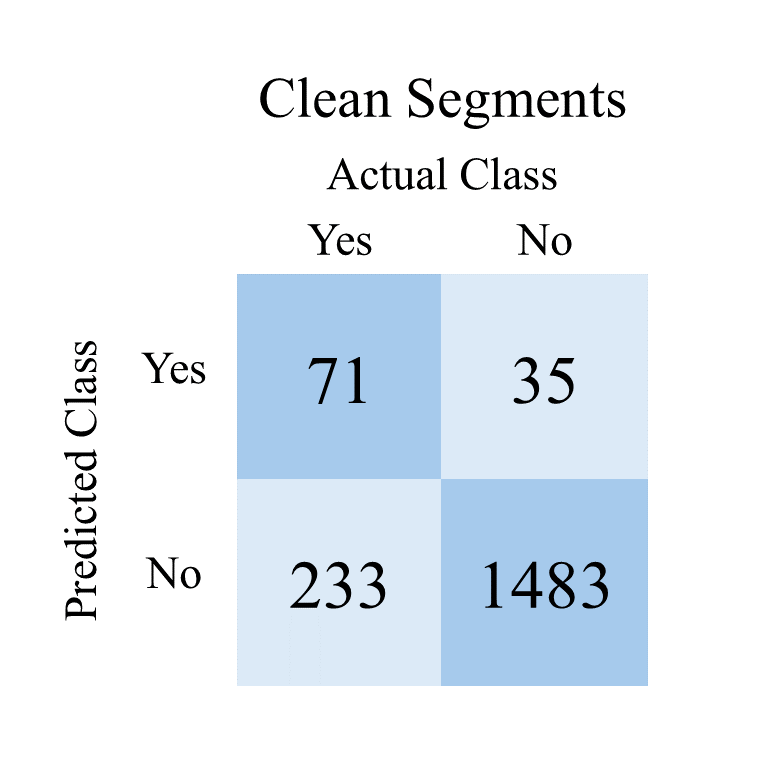}} \hfill
\captionbox{Confusion matrix of the prediction model using noisy segments and noisy audio. \label{fig:noisy_cm}} {
  \includegraphics[width=0.4\textwidth]{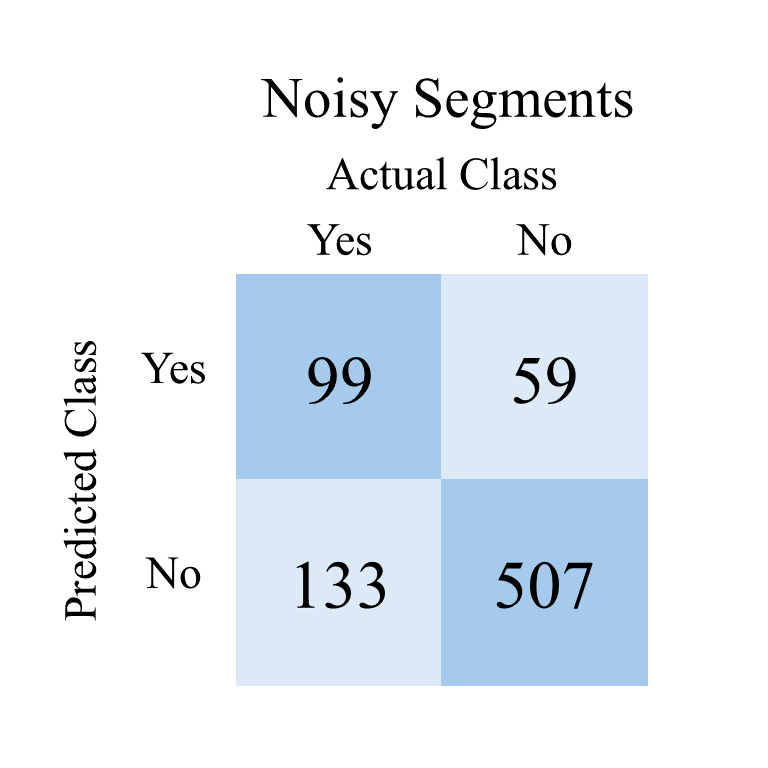}}
\end{figure*}

Table~\ref{tab:f1-noisy-noisy} shows results when noise from all 10 groups is added to the pipeline, both at the segmentation and transcription levels, which replicates the expected conditions in a real-world scenario where both automated segmentation and transcription are being used. Interruption detection still performs well but is slightly exceeded by Confirms Understanding, even when using acoustic features only.
In Figures~\ref{fig:clean_cm} and~\ref{fig:noisy_cm} we can also see that the distribution of labels from the clean segments has changed when we move into the noisy segments. Notably, noisy segments are typically longer because there are less periods of silence, leading to fewer overall segments and reduced granularity. The same phenomenon occurs between the oracle and clean segments. Thus, the total number of segments changes depending on the segmentation method used.
\begin{table*}
\centering
\caption{Average F1 score (standard deviation in parenthesis) using noisy segments and noisy audio.}
\setlength{\tabcolsep}{6pt} 
\begin{tabular}{rcccc}
\toprule
& \textbf{Interrupts} & \textbf{\begin{tabular}[c]{@{}l@{}}Confirms\\ Understanding\end{tabular}} & \textbf{\begin{tabular}[c]{@{}l@{}}Monitors\\ Results\end{tabular}} & \textbf{Responds} \\ \midrule
\textbf{Textual}     & .737 (.068) & .760 (.067) & .659 (.070) & .688 (.056) \\
\textbf{Acoustic}   & .746 (.094) & .753 (.060) & .652 (.082) & .703 (.051) \\
\textbf{Textual+Acoustic} & .751 (.089) & .768 (.066) & .676 (.103) & .699 (.045) \\
\hline
\end{tabular}
\label{tab:f1-noisy-noisy}
\end{table*}


\subsubsection{Experiment 3: Noisy Audio, Clean Segments}

Results from the final experiment, in which we used the timestamps from the clean segmented data to create segments from the noisy audio segments, are shown in Table~\ref{tab:f1-clean-noisy}. This experiment allowed us to explore the impact of noise on the feature quality independent of the segmentation. While interruption detection takes a further hit here, it is again the highest performing label, and detection of interruptions is robust to noise in the background compared to the other indicators, except for the \textit{responds} label. When ablating feature-wise, textual and acoustic features in interruption detection are both robust to background noise, with acoustic features being slightly more robust on average. 
 
\begin{table*}[ht!]
\centering
\caption{Average F1 score (standard deviation in parenthesis) using clean segments and noisy audio.}
\setlength{\tabcolsep}{6pt} 
\begin{tabular}{rcccc}
\toprule
& \textbf{Interrupts} & \textbf{\begin{tabular}[c]{@{}l@{}}Confirms\\ Understanding\end{tabular}} & \textbf{\begin{tabular}[c]{@{}l@{}}Monitors\\ Results\end{tabular}} & \textbf{Responds} \\ \midrule
\textbf{Textual}     & .782 (.052) & .749 (.063) & .720 (.095) & .769 (.048) \\
\textbf{Acoustic}   & .785 (.047) & .755 (.064) & .711 (.095) & .776 (.046) \\
\textbf{Textual+Acoustic} & .792 (.056) & .752 (.056) & .728 (.096) & .773 (.050) \\
\hline
\end{tabular}
\label{tab:f1-clean-noisy}
\end{table*}

\subsubsection{Acoustic Feature Analysis}
Following Experiments 1–3, we analyzed the best-performing acoustic models to identify which features most strongly influenced interruption detection. For each of the conditions (clean segments and clean audio, noisy segments and noisy audio, clean segments and noisy audio), we fit the best model to the entire dataset and used scikit-learn's feature importance using feature permutation. Table \ref{tab:ablation} shows the top three features for each experiment; a dash (–) indicates the feature did not rank among the top three for that experiment. We note that loudness features, such as median loudness, are consistently important; however, in the clean segments and noisy audio condition, features such as shape (MFCC3) and standard deviation of length of voices were also important.

\begin{table}[ht!]
\centering
\caption{Permutation feature importance for the best performing acoustic models in each experiment.}
\setlength{\tabcolsep}{6pt} 
\begin{tabular}{lccc}
\toprule
\textbf{Acoustic Feature} & \textbf{\begin{tabular}[c]{@{}l@{}}Clean-Clean \\      Importance\end{tabular}} & \textbf{\begin{tabular}[c]{@{}l@{}}Noisy-Noisy\\      Importance\end{tabular}} & \textbf{\begin{tabular}[c]{@{}l@{}}Clean-Noisy \\      Importance\end{tabular}} \\ \midrule
\textbf{Average F1 Amplitude} & .0175 & - & - \\
\textbf{Average Loudness} & - & \textbf{.0008} & - \\
\textbf{\begin{tabular}[c]{@{}l@{}}Range of 20th-80th \\\hspace{5mm} Percentile of Loudness\end{tabular}} & - & .0001 & \textbf{.0622} \\
\textbf{Median Loudness} & \textbf{.0558} & .0004 & - \\
\textbf{\begin{tabular}[c]{@{}l@{}}80th Percentile of \\\hspace{5mm} Loudness\end{tabular}} & .0153 & - & - \\
\textbf{Average MFCC3} & - & - & .0324 \\
\textbf{\begin{tabular}[c]{@{}l@{}}Standard Deviation of \\\hspace{5mm} Length of Voices\end{tabular}} & - & - & .0509 \\ \bottomrule
\end{tabular}
\label{tab:ablation}
\end{table}

\section{Discussion}
\label{sec:disc}

The heuristic baselines and random forest results over clean data indicate that the prevalent definition of interruptions as overlapping speech is a sufficient, though imperfect, indicator of interruptions in ideal conditions. However, as shown in Figure 4, the performance of this naive approach plummets when noise from concurrent collaborating groups overlap --- this is similar to the noise we would see in a classroom. One reason is that assuming overlapping speech equates to interruption means that if two utterances from two different groups overlap, it will be labeled an interruption, even though the speakers of the respective utterances are not speaking over each other intentionally. Even when only 2 other groups are present, the heuristic baseline labels almost every utterance as an interruption and ends up worse than chance prediction. Additionally, the presence of overlapping speech impacts automated segmentation and leads to lengthier and more coarse-grained utterances, which can be seen in Table~\ref{tab:utt-len}.

\begin{table*}[ht!]
\caption{Average utterance length of segments corresponding to each indicator type.}
\resizebox{\textwidth}{!}{
\begin{tabular}{lcc|cc|cc}
\hline
 & \multicolumn{2}{c|}{\textbf{Oracle Segments}} & \multicolumn{2}{c|}{\textbf{Clean Segments}} & \multicolumn{2}{c}{\textbf{Noisy Segments}} \\
\textbf{Indicator} & \textbf{Avg. Sec (SD)} & \textbf{Count} & \textbf{Avg. Sec (SD)} & \textbf{Count} & \textbf{Avg. Sec (SD)} & \textbf{Count} \\ \hline
\textbf{Interrupts} & 2.92 (16.52) & 378 & 6.57 (3.48) & 304 & 20.53 (14.38) & 232 \\
\textbf{\begin{tabular}[c]{@{}l@{}}Confirms\\ Understanding\end{tabular}} & 3.85 (3.82) & 326 & 5.74 (3.51) & 300 & 20.76 (14.58) & 228 \\
\textbf{\begin{tabular}[c]{@{}l@{}}Monitors\\ Results\end{tabular}} & 3.39 (15.47) & 423 & 5.01 (3.12) & 353 & 16.73 (13.71) & 287 \\
\textbf{Responds} & 2.02 (2.28) & 301 & 4.92 (2.97) & 265 & 19.57 (13.96) & 213 \\ \hline
\end{tabular}}
\label{tab:utt-len}
\end{table*}

\begin{figure}[t]
\centering
  \includegraphics[width=.8\columnwidth]{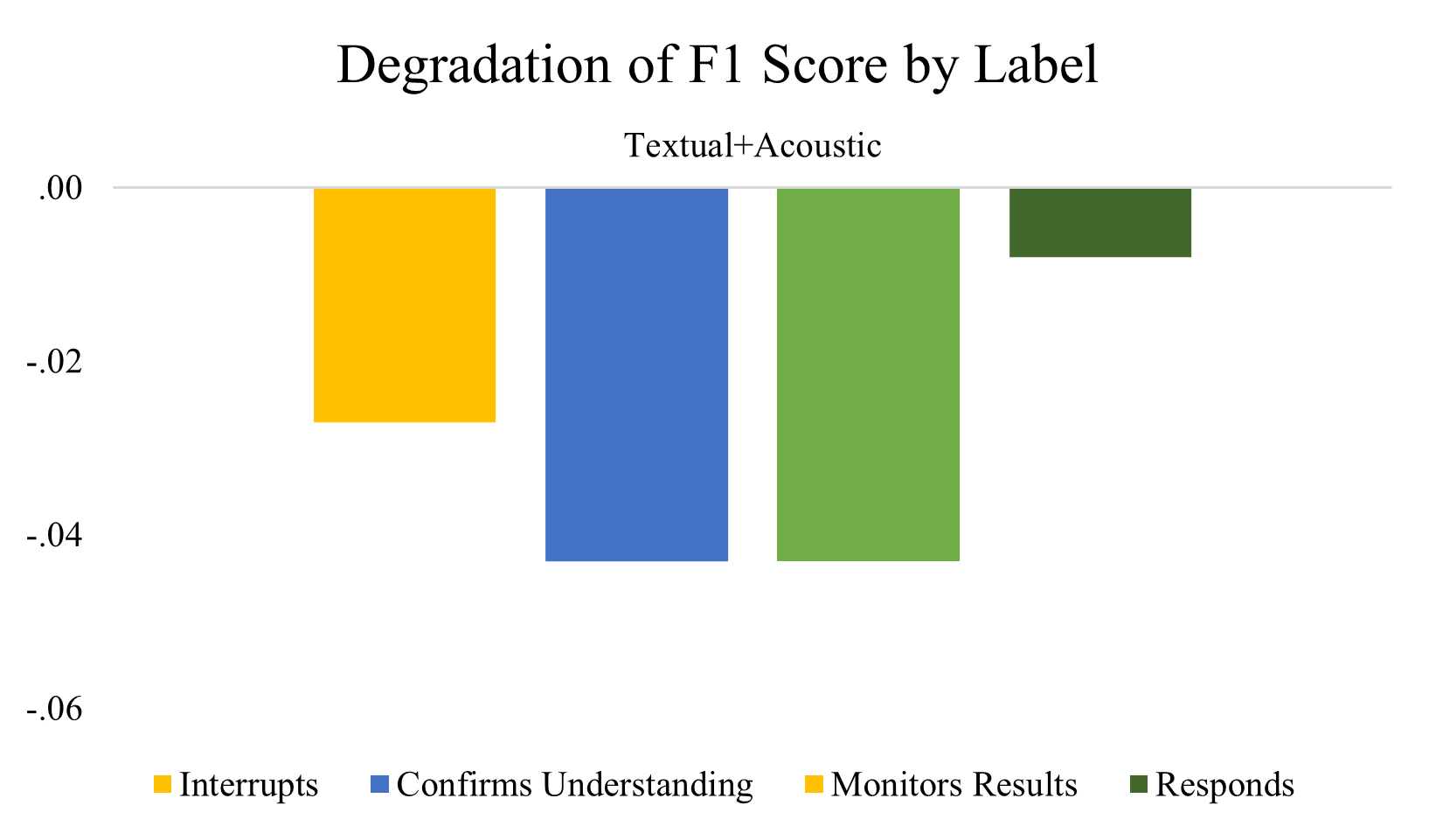}
  \caption{Degradation in F1 score by label between the clean (Experiment 1) and noisy audio (Experiment 3) when using the clean segmentation timestamps.}
  \label{fig:labelchart}
\end{figure}

Experiment 3---in which we applied the timestamps from the clean audio segmentation to the noisy audio transcriptions---revealed an interesting characteristic of the acoustic features. These features were successful in detecting interruptions \textit{and} they were robust to noise. When comparing the impact of noise on the other labels, this phenomenon revealed that interruptions still had the best detection performance when using acoustic features, especially in conjunction with textual features. This suggests that the acoustic features were able to provide necessary information for a model to learn \textit{main-speaker overlap}. By analyzing the acoustic features further, we see that loudness features are particularly important; however, when we add noise to clean segmentation, the shape (MFCC3) of the acoustics - such as a sudden shift - are indicative of interruptions.
This reflects the importance of including these features for model robustness to noisy settings. Where textual features may manifest as unexpected shifts in semantic content, acoustic features show changes in loudness, shape, or voice length, which are particularly important to detecting interruptions.

\begin{figure}[t]
\centering
  \includegraphics[width=.8\columnwidth]{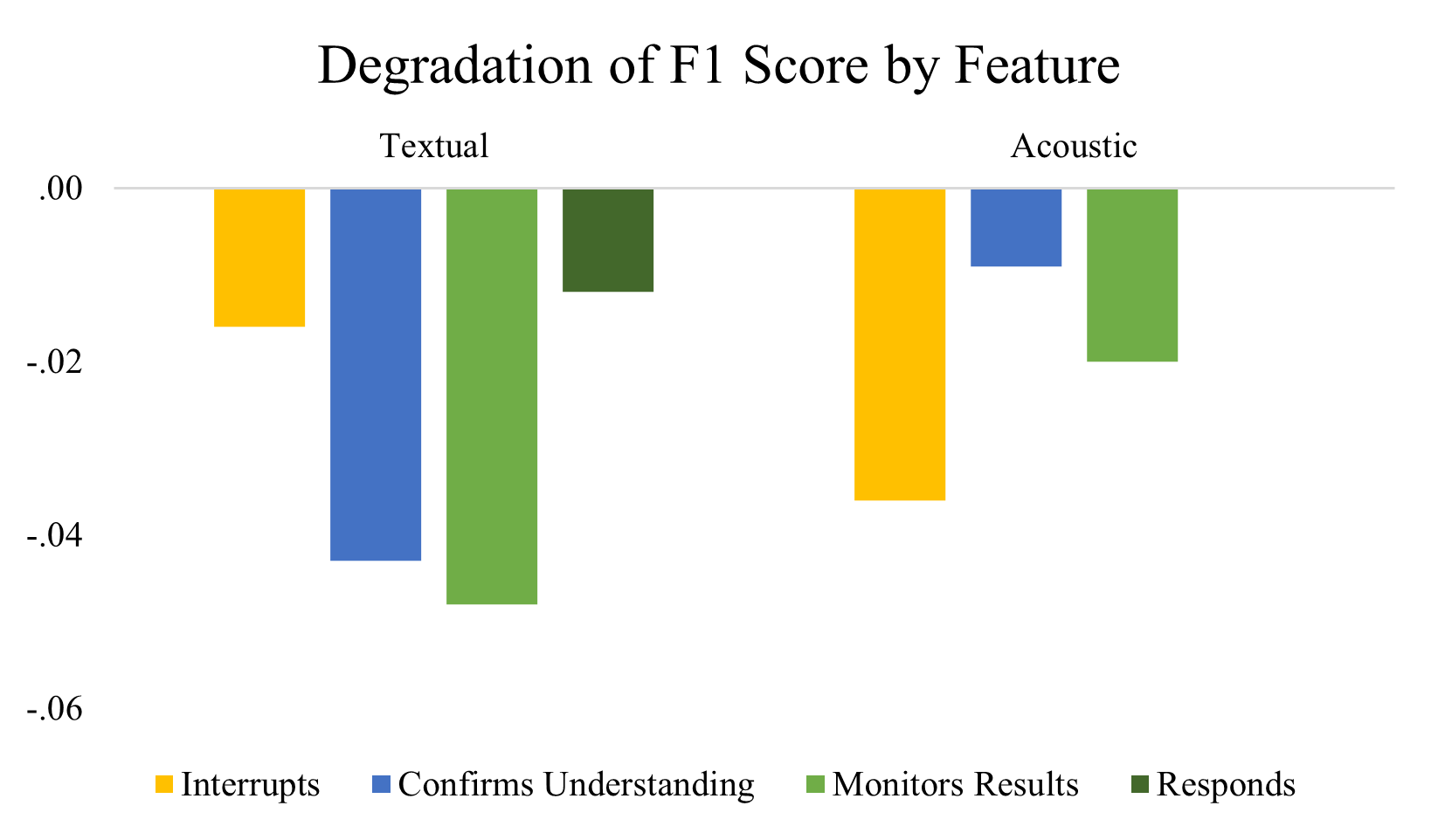}
  \caption{Degradation in F1 score by feature between the clean (Experiment 1) and noisy audio (Experiment 3) when using the clean segmentation timestamps.}
  \label{fig:featurechart}
\end{figure}

Figure \ref{fig:labelchart} shows that interruptions, which was the best-performing label initially, had a lower drop in performance than other labels---with the exception of responds---between Experiment 1 and Experiment 3 when using a combination of textual and acoustic features. 
Additionally, Figure \ref{fig:featurechart} reveals that between Experiment 1 and Experiment 3, the textual features suffered a notably higher degradation than the acoustic features on average.

\section{Conclusion and Future Work}
In this paper, we demonstrate the feasibility of detecting interruptions in natural human speech, even with background conversation noise. Using a modular automatic pipeline, we analyzed the impact of noise on interruption detection and feature importance, showing that interruptions remain detectable when combining acoustic and textual features. In contrast, other conversational labels proved more difficult to detect, highlighting the uniqueness of the interruption detection task.
Our goal was to evaluate whether a fully automated system could perform in both clean and noisy settings without relying on simple speech overlap. By introducing noise and comparing results, we confirmed that the model learns meaningful patterns beyond overlapping speech—a critical requirement for multi-group environments like classrooms.
We found that even under noisy conditions, acoustic and textual features retained signal, supporting the feasibility of applying this approach in real-world classroom settings. This work lays the groundwork for interruption detection systems that reduce teachers' cognitive load by helping them quickly identify groups needing support.
By decoupling interruption detection from simple overlap, our findings address a key limitation in prior work and demonstrate the promise of this method for authentic educational environments.

While this study uses a limited feature set, future work will expand it. Speaker diarization and overlapping speech detection --- focused within groups --- could enhance interruption detection. Visual speaker detection could further help by clustering nearby speakers into groups and tracking turn-taking. Future work should also distinguish between cooperative and competitive interruptions, which differ in features like pitch~\cite{yang_visualizing_2003}, enabling more informative alerts for teachers. Finally, these experiments were conducted in simulated multi-group settings, which may differ from real classroom environments.
Applications of this work include tracking and evaluating collaborative groups in classroom settings. Being able to focus on one conversation out of many using multiple channels is also reminiscent of the cocktail party effect seen in humans and could serve as a downstream task for evaluating systems focusing on one conversation in the presence of many.

\section*{Acknowledgments}
This material is based in part upon work supported by Other Transaction award HR00112490377 from the U.S. Defense Advanced Research Projects Agency (DARPA) Friction for Accountability in Conversational Transactions (FACT) program, and by the National Science Foundation (NSF) under a subcontract to Colorado State University on award DRL 2019805 (Institute for Student-AI Teaming). Approved for public release, distribution unlimited. Views expressed herein do not reflect the policy or position of, the Department of Defense, the National Science Foundation, or the U.S. Government. All errors are the responsibility of the authors. Our thanks also go out to the anonymous reviewers whose feedback helped improve the final copy of this paper.




\bibliographystyle{splncs04}
\bibliography{bib}

\end{document}